%% file: main.tex
\documentclass[10pt,twocolumn,letterpaper]{article}

\usepackage[pagenumbers]{cvpr} % To force page numbers, e.g. for an arXiv version


\definecolor{cvprblue}{rgb}{0.21,0.49,0.74}
\usepackage[pagebackref,breaklinks,colorlinks,allcolors=cvprblue]{hyperref}
\usepackage{comment}
\usepackage{multirow}
\usepackage{boldline}
\usepackage{makecell} 
\usepackage{arydshln}
\usepackage{algorithm}
\usepackage{algpseudocode}
\usepackage{xcolor}
\usepackage{float}
\usepackage{placeins}
\definecolor{codegreen}{rgb}{0,0.4,0}
\usepackage{url}

\def\paperID{8481} % *** Enter the Paper ID here
\def\confName{CVPR}
\def\confYear{2025}

\title{Urban Air Temperature Prediction using Conditional Diffusion Models}

\author{Siyang Dai, Jun Liu, Ngai-Man Cheung\\
Singapore University of Technology and Design\\
}

\begin{document}
\maketitle
\input{sec/0_abstract}    
\input{sec/1_intro}

\input{sec/2_related_work}
\input{sec/3_dataset}
\input{sec/4_method}
\input{sec/5_experiment}
\input{sec/6_application}
\input{sec/7_conclusion}

\clearpage
% \FloatBarrier

{
    \small
    \bibliographystyle{ieeenat_fullname}
    \bibliography{main}
}


\end{document}

%% file: sec/0_abstract.tex
\begin{abstract}
Urbanization as a global trend has led to many environmental challenges, including the urban heat island (UHI) effect. 
The increase in temperature has a significant impact on the well-being of urban residents.
Air temperature ($T_a$) at 2m above the surface is a key indicator of the UHI effect.
How land use land cover (LULC) affects $T_a$ is a critical research question which requires high-resolution (HR) $T_a$ data at neighborhood scale.
However, weather stations providing $T_a$ measurements are sparsely distributed e.g. more than 10km apart; and numerical models are impractically slow and computationally expensive.
In this work, we propose a novel method to predict HR $T_a$ at 100m ground separation distance (gsd) using land surface temperature (LST) and other LULC related features which can be easily obtained from satellite imagery.
Our method leverages diffusion models for the first time to generate accurate and visually realistic HR $T_a$ maps, which outperforms prior methods.
We pave the way for meteorological research using computer vision techniques by providing a dataset of an extended spatial and temporal coverage, and a high spatial resolution  
as a benchmark for future research.
Furthermore, we show that our model can be applied to urban planning by simulating the impact of different urban designs on $T_a$.
\end{abstract}

%% file: sec/1_intro.tex
\section{Introduction}
\label{sec:intro}

\begin{figure}[ht]
    \centering
    \includegraphics[width=\columnwidth]{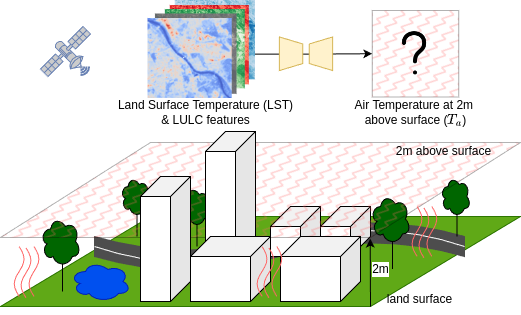}
    \caption{Task of HR $T_a$ prediction given LST and LULC features derived from satellite imagery.}
    \label{fig:task_summary}
\end{figure}

\begin{figure*}[htbp]
    \centering
    \includegraphics[width=\textwidth]{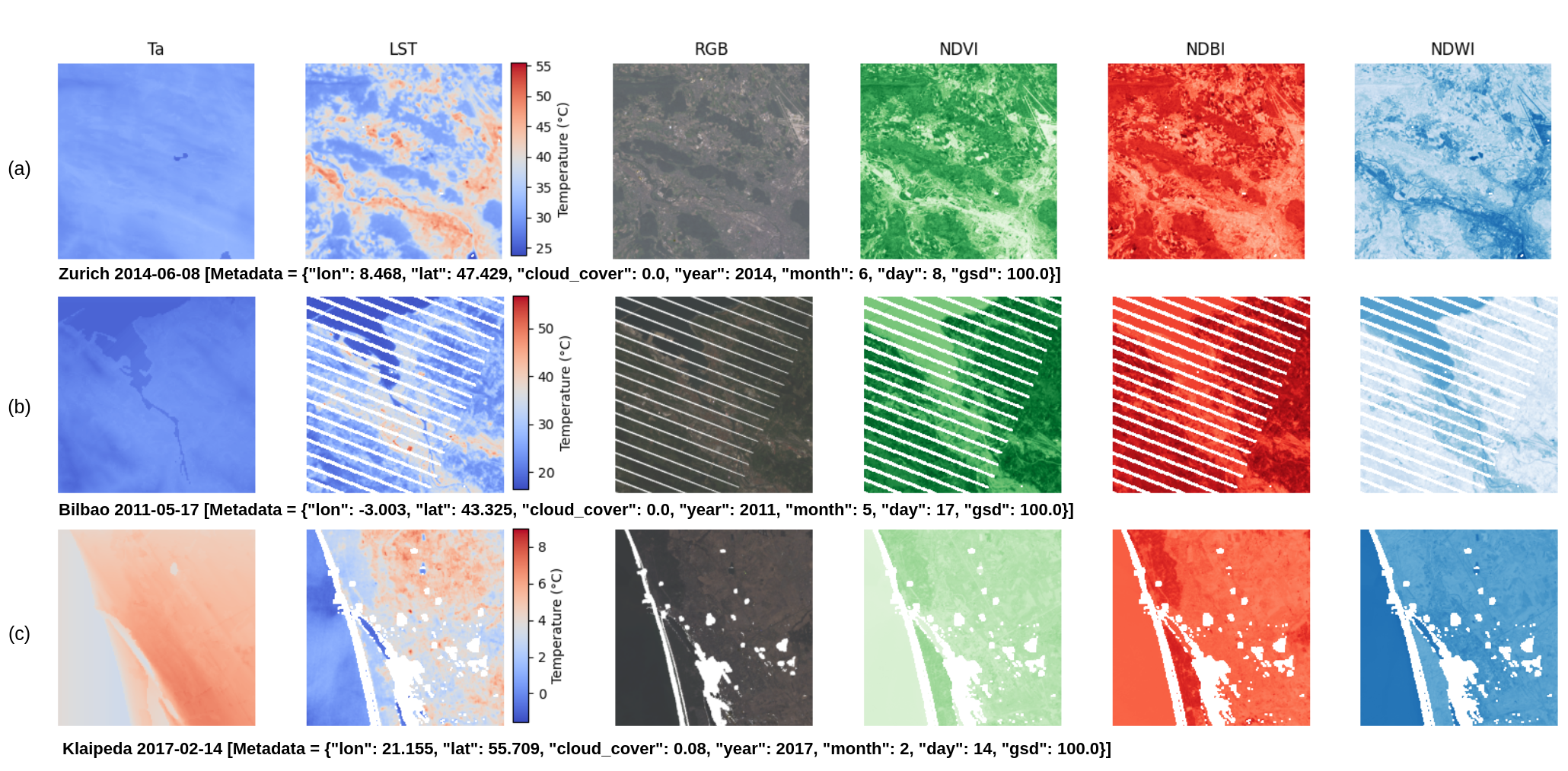}
    \caption{Illustration of LSTAT-20K dataset. 
    Each row represents a paired-up sample of $T_a$, LST, RGB, NDVI, NDBI and NDWI images and metadata, with $T_a$ and LST plotted on a common scale for visual comparison
    ($T_a$ is generally more uniform and has a narrower range compared to LST.).
    Samples (b) and (c) illustrate scan line failures and clouds on the satellite images respectively.
    }
    \label{fig:illustrate_dataset}
\end{figure*}

Air temperature as a key climate variable is a critical factor in urban planning, energy consumption, and human well-being.
To study the impact of urbanization, HR $T_a$ data at neighborhood scale is essential.
Weather stations can only provide highly sparse point measurements, which don't capture the fine-grained temperature patterns e.g. $T_a$ around buildings or trees.
Many studies on $T_a$ \cite{heGPRChinaTemp1kmHighresolutionMonthly2022,SpatialDownscalingERA5,wakjiraGriddedDaily2m2023} rely on the reanalysis dataset with super low spatial resolution. 
For example, ERA5 \cite{era5complete2024} has a gsd of 31km and the "land" variant \cite{era5land2024} has 9km gsd.
Our work is the first to provide a benchmark dataset for HR $T_a$ prediction at 100m gsd.
Researchers use numerical models to simulate HR $T_a$, but these models are computationally expensive due to the need of boundary conditions and forcing data etc.
Take the Envi-Met software \cite{envimet2024} as an example, 
a case study of microclimate's influence on vegetation with a small model (100x60x30 grids) 
takes about 164 hours with a Intel i7 CPU \cite{envimet2024casestudy}.

Prior works on $T_a$ prediction lack a common benchmark.
There is no common set of defined features, 
but frequently used are LST
and LULC related features.
The $T_a$ resolution that can be predicted depends on the resolution of these input features.
The authors of \cite{chajaeiMachineLearningFramework2024} try to estimate $T_a$ at 5m gsd 
based on proprietary LiDAR data (used for deriving urban geometry) of very high density.
However, the LiDAR and land use data as inputs does not align with the study period, during which the urban geometry may have changed significantly.
Similar for other input features, it's very challenging or even impossible to obtain feature data that is both time-aligned and of HR (spatial and temporal).
In this work, we create a benchmark dataset that not only 
contains the highest-resolution $T_a$ that is publicly available,
but we also ensure that all images in the dataset are time-aligned to the nearest hour.
The details of the dataset will be discussed in \cref{sec:dataset}.

We leverage diffusion models to predict $T_a$ maps from various input features among which LST is the most important.
We consider thermal conduction \cite{wiki_thermal_conduction,bobes-jesusAsphaltSolarCollectors2013},
which is the spread of heat energy within a single material or across materials that are in contact with each other,
as a diffusion process.
Specifically, the temperature gradient along the vertical direction from land surface to 2m above the surface
resembles the denoising diffusion process,
where we regard the noise removed at each time step as the incremental temperature change over height.
In addition, using diffusion models with conditional control can well preserve the spatial patterns in $T_a$ maps.

In summary, our contributions are as follows:
1. We create a benchmark dataset for HR $T_a$ prediction, which covers a wide spatial area with extended temporal span.
2. We propose a novel method to predict HR $T_a$ leveraging diffusion models and achieve state-of-the-art performance.
3. We demonstrate the potential of our method in urban planning by simulating the impact of different urban designs on $T_a$.

%% file: sec/2_related_work.tex
\section{Related work}
\label{sec:related_work}

%-------------------------------------------------------------------------
\subsection{Air temperature prediction}

Most prior works on air temperature prediction are case studies of a single city or within a country for a limited time duration.
The authors of \cite{BECHTEL201722} study the UHI characteristics of German cities using data from 2010 to 2012.
Other similar works conduct studies respectively in Sydney from 2019 to 2022 \cite{NASERIKIA2023167306}, in Amsterdam for 4 months in 2017 \cite{chajaeiMachineLearningFramework2024}, and in Hangzhou using 15 cloud free satellite images from 2006 to 2011 \cite{shengComparisonUrbanHeat2017}.
The majority of prior works focus on the thermal impact in summer times \cite{wangMachineLearningApplications2023}, leading to a lack of understanding of the seasonal temperature variations.
Recently, data-driven methods have been applied to prediction of urban air temperature.
The prior studies use mostly statistical modelling \cite{STRAUB2019100491,yiDevelopmentUrbanHighResolution2018,janatianStatisticalFrameworkEstimating2017a,shengComparisonUrbanHeat2017} and traditional machine learning models such as Random Forest \cite{venterCrowdsourcedAirTemperatures2021,venterHyperlocalMappingUrban2020,hutengsDownscalingLandSurface2016,eldesokyHighresolutionAirTemperature2021,chenHighresolutionMonitoringApproach2022a,chenIntegratingWeatherObservations2022,chenCombiningCityGMLFiles2020}, Multiple Linear Regression \cite{janatianStatisticalFrameworkEstimating2017a,chenStandardizingThermalContrast2021,caoCitySpatialTemporal2021a,bechtelSatelliteBasedMonitoring2017,alonsoNewApproachUnderstanding2020,alonsoIntegratingSatelliteDerivedData2019}, Gradient Boosting \cite{houghMultiresolutionAirTemperature2020,chajaeiMachineLearningFramework2024}, Support Vector Regression \cite{salcedo-sanzMonthlyPredictionAir2016,hoMappingMaximumUrban2014}, 
which don't take advantage of the spatial patterns in the data.
To our best knowledge, our work is the first to leverage diffusion models for HR $T_a$ generation, which can yield high-fidelity and spatially coherent $T_a$ maps.

%-------------------------------------------------------------------------
\subsection{Diffusion models} 
Diffusion models have achieved wide recognition in the computer vision community for their ability to generate high-quality images.
Not only are diffusion models used for text-to-image applications \cite{avrahamiBlendedDiffusionTextdriven2022,10204579,10030137,10205339,shiDragDiffusionHarnessingDiffusion2024}, 
but they have also been applied to various inverse tasks such as image inpainting \cite{lugmayrRePaintInpaintingUsing2022,zhuDenoisingDiffusionModels2023,xie2023smartbrush}, super-resolution \cite{yueResShiftEfficientDiffusion,wangSinSRDiffusionBasedImage2023}, and deblurring \cite{RefusionEnablingLargeSize}.
Recently, studies on dense prediction \cite{tian2024diffuse,nguyen2024dataset,leeExploitingDiffusionPrior2024}
have also started to leverage diffusion models.
Though diffusion models are stochastic in nature, methods \cite{lu2022dpm,Karras2022edm,song2020denoising} have been proposed to make them deterministic for dense prediction tasks.
One milestone work in constraining generated images is the ControlNet \cite{10377881}, which takes guidance from conditioning images.
Many works build on this idea to enable applications in different domains e.g. depth estimation \cite{yangDepthAnythingUnleashing2024}, remote sensing \cite{khannaDiffusionSatGenerativeFoundation2024},
image editing \cite{chenAnyDoorZeroshotObjectlevel2024}, and super-resolution \cite{yuScalingExcellencePracticing2024,linDiffBIRBlindImage2024}.
We are the first to apply diffusion models with ControlNet for HR $T_a$ prediction.

%% file: sec/3_dataset.tex
\section{Task and dataset}
\label{sec:dataset}

%-------------------------------------------------------------------------
\subsection{Task settings}
\label{sec:task_definition}

We introduce three task settings for HR $T_a$ prediction.

\textbf{Same resolution}. This is our baseline task that predicts $T_a$ given LST and LULC features 
(RGB, NDVI, NDBI, and NDWI).
Both target and input features are at the same resolution of 100m gsd,
i.e. $LST^{100m} + LULC^{100m}\rightarrow T_a^{100m}$.

\textbf{Super-resolution (SR)}. 
This setting applies when we have the more accessible low-resolution (LR) $T_a$ available, such as LR simulation.
We reduce the resolution of $T_a$ by a factor of 3
and use it as an additional condition to predict the original resolution $T_a$, 
i.e. $T_a^{300m} + LST^{100m} + LULC^{100m} \rightarrow T_a^{100m}$.

\textbf{Point SR}. We further reduce the $T_a$ to a few pixels, 
i.e. $T_a^{N pts} + LST^{100m} + LULC^{100m} \rightarrow T_a^{100m}$.
This simulates the point measurements from weather stations,
where $N$ is the number of stations within the $T_a$ map.
We use a grid of 10km x 10km to sample the original $T_a$ as the $N$ point measurements. 

On top of the two SR tasks, to push for highest resolution $T_a$ prediction,
we introduce the inferencing of $T_a$ at 30m gsd 
given the 100m gsd $T_a$.
As the ground truth for $T_a^{30m}$ is not available, 
we directly inference using the models trained in the SR and Point SR settings, i.e.
$T_a^{100m} + LST^{100m} + LULC^{30m} \rightarrow T_a^{30m}$.
The evaluation is performed on a downsampled version of the generated $T_a^{30m}$ images 
for comparison with original $T_a^{100m}$.

%-------------------------------------------------------------------------
\subsection{Dataset}

We create a benchmark dataset to facilitate future research in HR $T_a$ prediction.
To achieve this, we source the target HR $T_a$ from UrbClim\footnote{Generated using Copernicus Climate Change Service information, 2019. Neither the European Commission nor ECMWF is responsible for any use that may be made of the information or data it contains.}  \cite{Copernicus_Climate_Data},
and satellite imagery for LST and LULC features from Landsat \cite{usgs_landsat}.

\textbf{UrbClim}.
An urban climate model that produces high-fidelity climate variables 
verified against in-situ measurements \cite{UrbclimFastUrban,chajaeiMachineLearningFramework2024,garcia-diezAdvantagesUsingFast2016}.
The $T_a$ data provided supports UHI studies, 
but has not been previously combined with satellite imagery for a comprehensive CV benchmark.
As mentioned in \cref{sec:intro}, climate simulation is expensive and slow, 
but data-driven methods based on such a benchmark can significantly improve the efficiency.
Furthermore, with satellite imagery alone in the future, 
$T_a$ can be predicted without the need for running new simulations.
UrbClim covers a large spatial area of 100 European cities which exhibit a high variance in climate conditions, 
and an extended period from 2008 to 2017.

\textbf{Landsat} 
% \cite{usgs_landsat}.
A series of Earth observation satellites that provide multi-spectral images 
at 30m spatial resolution and thermal infrared band at 100m effective resolution, 
and a temporal resolution of 16 days.
We use Landsat's multi-spectral bands to derive LST, NDVI, NDBI, NDWI, and RGB images for the dataset detailed in \cref{sec:data_preparation}.

%-------------------------------------------------------------------------
\subsubsection{Data preparation and preprocessing}
\label{sec:data_preparation}

As first step to prepare the dataset, we use the cities in UrbClim to locate and download Landsat images.
Then to pair up $T_a$ and satellite data in location and time, we take the  spatial coverage of each city as the mask to crop the satellite images,
and use the timestamp of the satellite image to filter $T_a$ up to hourly precision.
Then we prepare the raw features 
from the multi-spectral bands.
For LST, we directly scale the thermal infrared band (tir) using
$lst = m*tir + a$, where $m$ and $a$ are the provided multiplier and bias. 
Then the indexes are computed using equations 
$ndvi = \frac{nir - red}{nir + red}$,
$ndbi = \frac{swir - nir}{swir + nir}$, and
$ndwi = \frac{nir - swir}{nir + swir}$,
where nir, swir are the near-infrared, and short-wave infrared bands respectively.
Note that $T_a$, LST, NDVI, NDBI, and NDWI are single-channel images.
The 3-channel RGB images are created by stacking the red, green, and blue bands.
Existing works \cite{chenIntegratingWeatherObservations2022,chenStandardizingThermalContrast2021} take LULC as a key feature for the impact of different urban fabrics on $T_a$.
LULC that aligns in time with $T_a$ is rarely available, sometimes even years apart, which deteriorates the prediction accuracy.
In our case, we innovatively use RGB images which inherently contain LULC information and are always synchronized with $T_a$.
Likewise, the NDVI, NDBI, and NDWI indexes are used to capture impact of the vegetation, built-up areas, and water bodies on $T_a$ respectively.
We also prepare the metadata for each pair of $T_a$ and input features including latitude, longitude, cloud cover, year, month, day and gsd.
We show an illustration of the dataset in \cref{fig:illustrate_dataset}.

For image preprocessing, depending on the city size from UrbClim 
where $T_a \in \mathbb{R}^{101 \times 101} \sim \mathbb{R}^{401 \times 401}$,
the satellite images are of $3.33\times$ larger in size, ranging from
$I_{sat} \in \mathbb{R}^{332 \times 332} \sim \mathbb{R}^{1334 \times 1334}$.
To align with the input size of the pretrained model, 
we partition the larger ($>\mathbb{R}^{512 \times 512}$) satellite images into $512 \times 512$ patches 
and process the paired $T_a$ images accordingly,
Since LST has an effective resolution of 100m gsd,
we downsample LST to the same size as $T_a$ using bi-linear interpolation.
This data preparation ensures no information loss during training.

One challenge in preparing Landsat satellite images is the invalid pixels due to the land-obstructing clouds and scan line failures with Landsat 7 \cite{usgs_landsat7_etm}.
Prior works \cite{samehAutomatedMappingUrban2022,shengComparisonUrbanHeat2017,RAO2023e18423} are inclined to select cloud-free images resulting in limited data.
We, on the other hand, accept all images up to a 20\% cloud cover and with scan line failures.
We then inpaint the invalid pixels using the same diffusion framework as the one used for $T_a$ prediction.
Details are provided in the supplementary material. 

%-------------------------------------------------------------------------
\subsubsection{Dataset statistics and splits}
\label{sec:dataset_stats}

Our dataset contains 100 cities in Europe, each with 10 years of data.
The total geospatial area covered by the dataset is 59,179 $km^2$, averaging 591 $km^2$ per city. 
The preprocessing results in a total of 20,410 samples each containing images for $T_a$, LST, NDVI, NDBI, NDWI, and RGB along with the metadata.
To this end, we name our dataset as \textit{\textbf{LSTAT-20K}}.
Temporal wise, the number of satellite images is limited by the Landsat revisit cycle and cloud cover requirement.
Nonetheless, we try to include as many images to cover all months for each city, 
although certain cities and months have fewer samples due to more cloudy days.
For example, continental cities have fewer samples than Mediterranean cities;
and winter months have fewer samples than summer months.
We show the distribution of the number of images per city and per month in the supplementary material.

We split the dataset into training and test sets using samples from years 2008 to 2015 and years 2016 to 2017 respectively, resulting a 3:1 ratio.
The intuition is to train on the historical data and predict on future data to evaluate the generalization of the model.

%% file: sec/4_method.tex
\section{Methodology}

\begin{figure*}[htbp]
    \centering
    \includegraphics[width=\textwidth]{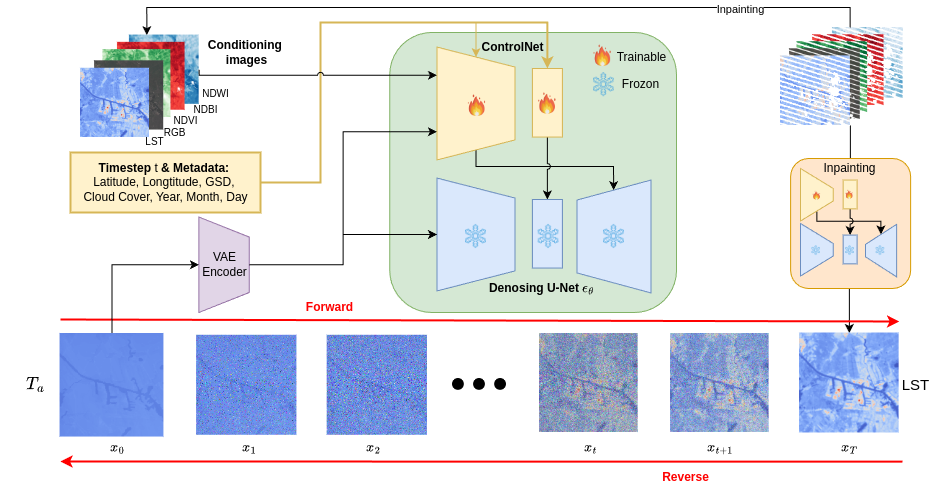}
    \caption{The pipeline of DiffTemp.
    The forward process adds noise to the target $T_a$ image until close to the LST image.
    The ControlNet consumes the target latent of $T_a$ and takes as conditioning images the satellite-derived LST, RGB and the index images and metadata.
    The denoising U-Net predicts the noise at each step given the target latent of $T_a$ and residuals from ControlNet's downsampling and middle blocks.
    The reverse process removes noise from the LST image to recover the target $T_a$ image.
    }
    \label{fig:pipeline}
\end{figure*}

We are motivated by the idea that the gradual temperature change along the vertical direction from LST to $T_a$ resembles the denoising diffusion process.
By treating each incremental temperature change $\delta T$ over a certain height $\delta h$ to the noise removed at each sampling step,
we consider the forward process to be adding noise to the target $T_a$ image, 
and the reverse process to be removing the noise from the LST image.
We first provide the background of diffusion models and ControlNet in \cref{sec:background}.
Then we describe our $T_a$ prediction method DiffTemp 
in \cref{sec:ta_prediction} 
and satellite images inpainting in the supplementary.

%-------------------------------------------------------------------------
\subsection{Background}
\label{sec:background}

\textbf{Diffusion models}. 
Diffusion models are a class of generative models 
by iteratively removing noise from a noise-perturbed image.
DDPM \cite{NEURIPS2020_4c5bcfec} models the forward diffusion process by a Markov chain in \cref{eq:diffusion_forward_direct}
where $\bar{\alpha_t}$ is the variance schedule and $\epsilon_t \sim N(0,I)$ is the noise.

\begin{equation}
    x_{t} = \sqrt{\bar{\alpha_t}} x_{0} + \sqrt{1-\bar{\alpha_t}} \epsilon_t, \quad t \in [1, T]
    \label{eq:diffusion_forward_direct}
\end{equation}

The schedule is designed to weight $x_0$ more at the beginning and $\epsilon$ more at the end,
so the final $x_T$ is close to pure Gaussian noise.
Recovering the target data $x_0 \sim q(x_0)$ is by the reverse process in \cref{eq:diffusion_reverse},
where $p(x_{t-1}|x_t) \sim N(x_{t-1}; \mu_{\theta}(x_t, t), \Sigma_{\theta}(x_t,t))$.
The diffusion network $\epsilon_{\theta}$ that predicts the noise to be added at each step,
is optimized with the objective $\mathbb{E}_{x_0 \sim q(x_0), \epsilon \sim N(0,I)} \left\| \epsilon-\epsilon_{\theta} (x_t,t) \right\|_2^2$.
In this work, we follow \cite{10377881} to use the Stable Diffusion model \cite{Rombach_2022_CVPR},
which performs the diffusion process in the latent space using the variational autoencoder (VAE).

\begin{equation}
    p(x_{0:T}) = p(x_T) \prod_{t=1}^{T} p(x_{t-1}|x_t) 
    \label{eq:diffusion_reverse}
\end{equation}

\textbf{ControlNet}.
ControlNet \cite{10377881} is a conditional diffusion model that controls the generation process with  condition inputs such as depth or segmentation images.
By reusing a trainable copy of the pre-trained layers of the original model and 
connecting back via zero-convolution layers,
ControlNet can accurately drive Stable Diffusion to generate controlled images.
We choose to leverage ControlNet for our task to preserve the spatial patterns in the input features.

%-------------------------------------------------------------------------
\subsection{Air temperature prediction}
\label{sec:ta_prediction}
The model pipeline for HR $T_a$ prediction is shown in \cref{fig:pipeline}.
We use a combination of Stable Diffusion and ControlNet framework as in \cite{10377881}.
In summary, the main innovations include replacing the original noise schedule with a deterministic one to improve $T_a$ prediction accuracy, 
and incorporating metadata into the conditional inputs to enhance guidance based on cues such as location and time.
We provide the details in the following.

\textbf{Noise scheduling}.
Typical text-to-image (T2I) generations are stochastic,
so they use \cref{eq:diffusion_forward_direct} to diffuse  the target until Gaussian noise.
We observe that by denoising from pure Gaussian noise, the generated temperature values are far from the target $T_a$.
Motivated by the idea that thermal conduction process from LST to $T_a$ resembles the denoising diffusion process,
and inspired by scheduling ideas in \cite{heitzIterative$a$deBlendingMinimalist2023,leeExploitingDiffusionPrior2024,yueResShiftEfficientDiffusion},
instead of diffusing and denoising the target image to and from pure noise, we use a scheduling strategy that takes advantage of the key feature LST.
Specifically, we gradually add noise to the target $T_a$ until close to LST in the forward diffusion process, 
then denoise from LST to the target $T_a$ in the reverse process.
The forward process now becomes \cref{eq:diffusion_forward_new},
where $x_0$ is the $T_a$ image and $y$ is the LST image,
both being used for all time steps.
This way the model can learn the temperature patterns from the LST and predict the $T_a$ values that are close to the target.
Similar to the noise scheduling in \cref{eq:diffusion_forward_direct}, where $x_t$ is more close to the target $x_0$ at the beginning and more close to pure noise $\epsilon$ towards the end,
the new scheduling strategy in \cref{eq:diffusion_forward_new} weights the target $x_0$ ($T_a$) more at the beginning and $y$ (LST) more towards the end, 
as $\bar{\alpha_t}$ decreases when the diffusion timestep $t$ increases.

\begin{equation}
    x_{t} = \sqrt{\bar{\alpha_t}} x_{0} + \sqrt{1-\bar{\alpha_t}} y, \quad t \in [1, T]
    \label{eq:diffusion_forward_new}
\end{equation}

\textbf{Metadata}. 
Compared to natural images, bird-eye-view images like satellite and $T_a$ images have their unique characteristics, such as the geographical location (latitude and longitude) and the temporal information, 
which highly influence the temperature patterns and absolute values.
We construct metadata of the conditioning images and feed them to the ControlNet to guide the generation process like in \cite{khannaDiffusionSatGenerativeFoundation2024}.
Specifically, 
the metadata embeddings are added to the time step $t$ embedding as the input to the ControlNet.
Apart from the influence of aforementioned time and location on $T_a$, as we split the dataset by different years for the training and test sets, 
we can assess the model's generalization during inference since the model has never seen the test years' data during training.
The gsd helps the model to understand the spatial scale of the images,
and cloud cover hints the model about the missing pixels in the satellite images.

We illustrate the pipeline of HR $T_a$ prediction in \cref{fig:pipeline}.
Specifically, the target image $x_0 \in \mathbb{R}^{H \times W \times 3}$ is encoded to the latent space $z_0 \in \mathbb{R}^{h \times w \times c}$ by the VAE encoder, 
where $(h,w)=(H/8,W/8)$ and c is the number of latent channels.
The LST image $y$ is encoded to $z_y$ in the same way.
The actual diffusion process is performed in the latent space by simply replacing $x$ with $z$ and $y$ with $z_y$ in \cref{eq:diffusion_forward_new}.
The latent $z_t$ serves as input to both the ControlNet and the U-Net.
The ControlNet also takes the concatenated
LST and LULC images as conditions.
The metadata and time step $t$ embeddings are added then input to the downsampling and middle blocks of the ControlNet.
The residuals from ControlNet then guide the U-Net to predict at each step.
Instead of predicting noise, we find using v-prediction \cite{salimansProgressiveDistillationFast2022} leads to better performance in our task.
We use \cref{eq:loss_v_pred} as our objective to train the ControlNet,
where $v=\sqrt{\bar{\alpha_t}} z_y - \sqrt{1-\bar{\alpha_t}} z_t$ is the velocity term with $z_y$ as the LST latent,
and $v_{\theta}$ represents the U-Net and $c$ is the residuals from ControlNet.

\begin{equation}
    \mathcal{L}_{\text{mse}} = \mathbb{E}_{z_0, z_y} \left\| v-v_{\theta} (z_t, t, c) \right\|_2^2
    \label{eq:loss_v_pred}
\end{equation}

The reverse process is given in \cref{eq:diffusion_reverse_new},
where $z_T$ is the LST latent encoded by the VAE encoder i.e. $z_T = vae.enc(LST)$, 
and the denoised $\hat{z_0}$ latent is expected to predict the desired $z_0 = vae.enc(T_a)$.
The output $\hat{z_0}$ is decoded by the VAE decoder to obtain the final $T_a$ image i.e. $T_a = vae.dec(\hat{z_0})$.
Note that the VAE decoder is omitted in the pipeline \cref{fig:pipeline} for tidy look.

\begin{equation}
\begin{split}
    z_{t-1} = \sqrt{\bar{\alpha}_{t-1}} (\sqrt{\bar{\alpha_t}} z_t - \sqrt{1-\bar{\alpha_t}} v_{\theta} (z_t, t, c)) \\ 
    + \sqrt{1-\bar{\alpha}_{t-1}} z_y, 
    \label{eq:diffusion_reverse_new}
\end{split}
\end{equation}

%% file: sec/5_experiment.tex
\section{Experiments}
\label{sec:experiments}

We present the experimental analysis of our proposed method DiffTemp for HR $T_a$ prediction on our dataset LSTAT-20K.
We conduct extensive experiments to evaluate and compare the performance of DiffTemp with prior methods.
We also conduct an ablation study to analyze 
the impact of the noise scheduling strategy in the DiffTemp model.

%-------------------------------------------------------------------------
\subsection{Implementation details}
\textbf{Data resolution.}
(1) Same resolution task.
As mentioned in \cref{sec:data_preparation}, 
the LULC images have $3.33\times$ higher resolution (30m) than $T_a$ and LST images (100m).
To be specific, within a sample we have NDVI, NDBI, NDWI, and RGB images at a maximum of $512 \times 512$ resolution, 
and LST and $T_a$ images at $0.3\times$ resolution of the satellite images, i.e. $154 \times 154$ at maximum.
During training, we downsample the LULC images to the size of $T_a$ and LST  for alignment of the feature resolution.
(2) SR and Point SR tasks.
For the additional condition, 
we downsample $T_a$ to $300m$ resolution, i.e.
$52 \times 52$ at maximum, for the SR task.
For the Point SR task, we sample $N$ points from the $T_a$ images as point measurements.
We use the same sizes for the other images as in the same resolution task.
Finally, all images are transformed to $512 \times 512$ before feeding into the model.

\textbf{Training details.}
We employ the pretrained StableDiffusion \cite{Rombach_2022_CVPR} 
and the ControlNet architecture \cite{10377881},
and a U-Net that is pretrained on satellite images from \cite{khannaDiffusionSatGenerativeFoundation2024}.
We use the Adam optimizer with a learning rate of 5e-5 and a batch size of 4,
and update the model weights for 300k training steps.
More specifically, we fine-tune for each of the 100 cities in the dataset for 3k steps.
 
%-------------------------------------------------------------------------
\textbf{Compared methods.}
We set n\_estimators=100, max\_depth=10 for both Random Forest and Gradient Boosting, 
and alpha=0.1 for Linear Regression, 
and hidden\_layer\_sizes=(64,128,64), lr=0.001 for MLP.
 
%-------------------------------------------------------------------------
\textbf{Evaluation metrics.}
We evaluate the performance of the $T_a$ prediction model using mean absolute error (MAE) and 
root mean squared error (RMSE) in degree celcius following the common practice in prior works \cite{BECHTEL201722,zhuEstimationDailyMaximum2013,NASERIKIA2023167306,wangMachineLearningApplications2023}.
In addition, we involve a new metric Structural Similarity Index (SSIM) \cite{1284395} to evaluate 
the spatial patterns preserved in the predicted $T_a$ maps.

%-------------------------------------------------------------------------
\subsection{Experimental results}
We show the results for the three task settings in this section.
As there is no public benchmark dataset for HR $T_a$ prediction,
we compare our method with the frequently used methods in prior works by the same-resolution task setting.
We select 4 representative methods for comparison:
Random Forest (RF), Gradient Boosting Decision Tree (GBDT), Linear Regression (LR), and Multi-layer Perceptron (MLP),
of which RF and GBDT are the best performing methods in most prior works.
In order to make a fair comparison, we utilize the same data features for all methods.
Specifically, we use LST, NDVI, NDBI, NDWI, and RGB as well as the metadata as input features and $T_a$ as target for all methods.
Since the machine learning (ML) models require tabular data, 
we build a dataset that regards a pixel as a data entry.
We train a separate model for each city in the dataset for better capturing the city-specific patterns.
To rule out the effect of the inpainted pixels, 
we use the raw data for training and 
evaluate only on the valid pixels. 

\textbf{Same resolution}. We show in \cref{tab:results_compare} the performance of all the methods on LSTAT-20K dataset.
It can be seen that our method DiffTemp outperforms the prior methods by a large margin, 
especially in terms of SSIM.
This indicates that DiffTemp can better capture the spatial patterns in the $T_a$ maps,
in addition to the low physical errors by metrics MAE and RMSE.
It's worth noting that even though the Linear Regression model has lower MAE and RMSE than the RF and GB models,
it has the lowest SSIM score, which exposes the limitation of regression methods to only mimic the target distribution of temperature values without understanding the spatial patterns.

\begin{figure*}[htbp]
    \centering
    \includegraphics[width=\textwidth]{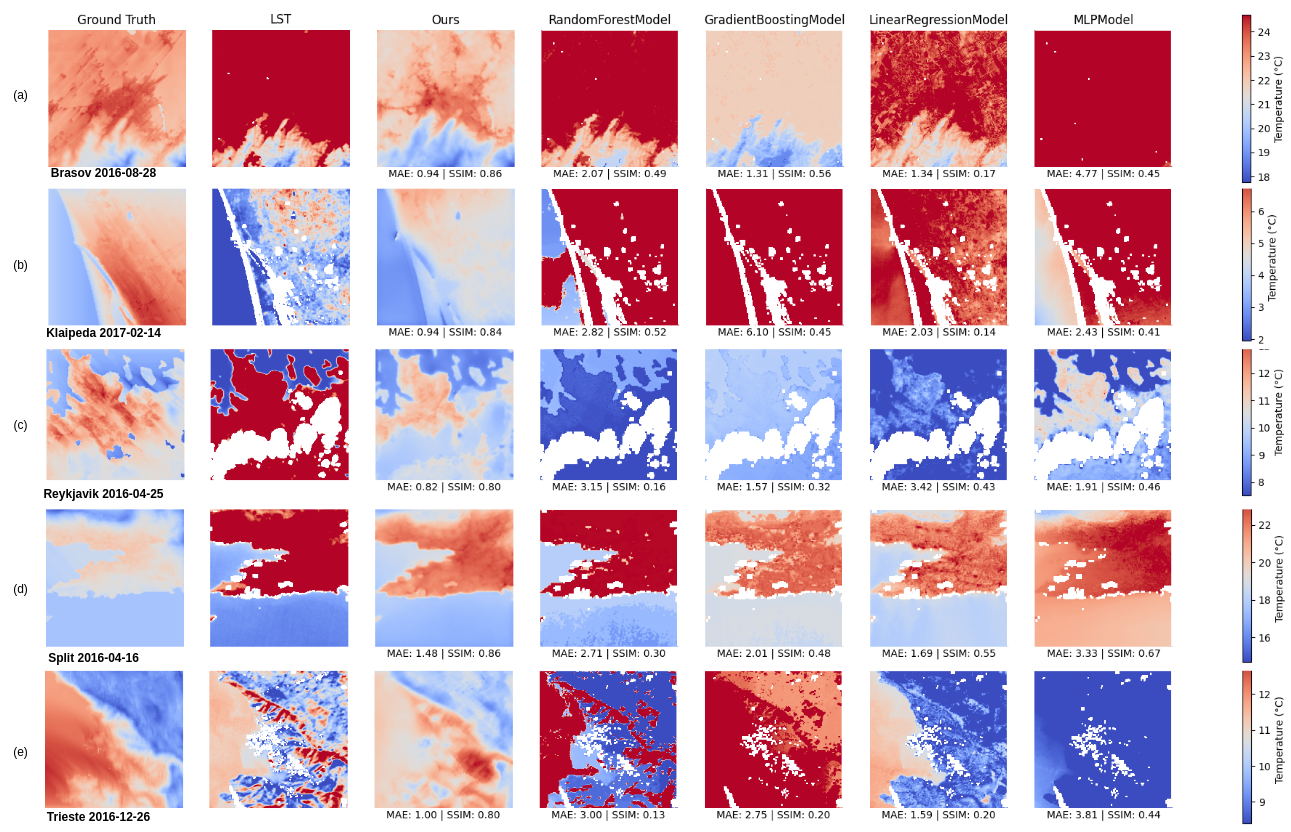}
    \caption{Qualitative comparison of the predicted $T_a$ maps by ours (DiffTemp) and the prior methods.}
    \label{fig:compare_results}
\end{figure*}

To further illustrate the effectiveness of DiffTemp, 
we show in \cref{fig:compare_results} the visual comparison of the predicted $T_a$ maps by DiffTemp and the prior methods.
It can be seen that the ML methods tend to follow the input LST features, 
which actually ranks the highest in the feature importance analysis of Random Forest model.
For example, in Sample (a), RF and Linear Regression models predict the $T_a$ maps that are very similar to the LST maps,
and same for RF model in Sample (d).
In Sample (c), the ML methods deviate a lot from the ground truth.
This indicates that ML models are not able to generalize well to all the samples.

\begin{table}
    \small
    \centering
    \begin{tabular}{cccc}
        % \hline
        \toprule
        \thead{Method} & \thead{RMSE$\downarrow$} & \thead{MAE$\downarrow$} & \thead{SSIM$\uparrow$} \\
        \midrule
        Random Forest & 3.72 & 3.45 & 0.32 \\
        Gradient Boosting & 3.44 & 3.25 & 0.36 \\
        Linear Regression & 2.84 & 2.63 & 0.23 \\
        MLP & 3.54 & 3.42 & 0.45 \\
        \midrule
        DiffTemp (ours) & \textbf{2.20} & \textbf{2.09} & \textbf{0.70} \\
        \bottomrule
    \end{tabular}
    \caption{Comparison of the prior methods with DiffTemp on LSTAT-20K.}
    \label{tab:results_compare}
\end{table}

\begin{table}
    \small
    \centering
    \begin{tabular}{cccc}
        % \hline
        \toprule
        \thead{Task} & \thead{RMSE$\downarrow$} & \thead{MAE$\downarrow$} & \thead{SSIM$\uparrow$} \\
        \midrule
        % $T_a^{300m} \rightarrow T_a^{100m}$ & 0.966 & 0.895 & 0.768 \\
        % $T_a^{30m}$ (inference only) & 1.764 & 1.523 & 0.719 \\
        Super-resolution (SR) & 0.920 & 0.849 & 0.779 \\
        $T_a^{30m}$ with SR & 1.950 & 1.704 & 0.710 \\
        \midrule
        % $T_a^{N pts} \rightarrow T_a^{100m}$ & 1.805 & 1.693 & 0.713 \\
        % $T_a^{30m}$ (inference only) & 2.370 & 2.149 & 0.656 \\
        Point SR & 1.707 & 1.588 & 0.724 \\
        $T_a^{30m}$ with Point SR & 2.469 & 2.197 & 0.664 \\
        \midrule
        Same resolution & 2.208 & 2.097 & 0.701 \\
        \bottomrule
    \end{tabular}
    \caption{Evaluation of the SR and Point SR tasks and the inference on $T_a^{30m}$ with the respective model trained.}
    \label{tab:results_sr_ta}
\end{table}

\textbf{SR and Point SR}. By using the same model architecture as in \cref{sec:ta_prediction} with the downsampled $T_a$ as an additional conditioning image,
we train and evaluate the model on the two task settings,
and inference $T_a^{30m}$ with the same model trained.
We show the experimental results in \cref{tab:results_sr_ta}.

As the results show, when incorporating either the downsampled $T_a$ or the point measurements as an additional condition,
the model performance improves compared to the baseline task.
The SR task achieves better performance than the Point SR task.
This is expected since the downsampled $T_a$ contains more spatial information than the point measurements.

The inference on $T_a^{30m}$ with the model trained on SR task performs better than Point SR.
This indicates the model better transfers SR knowledge trained from $300m \rightarrow 100m$ to the inference task $100m \rightarrow 30m$ than Point SR.
However, even with point measurements alone, we can derive a meaningful $T_a$ map with reasonable accuracy and perceptual quality.
This enables future application incorporating weather station measurements.

\begin{table}
    \small
    \centering
    \begin{tabular}{cccc}
        % \hline
        \toprule
        \thead{Method} & \thead{RMSE$\downarrow$} & \thead{MAE$\downarrow$} & \thead{SSIM$\uparrow$} \\
        \midrule
        Pure noise scheduling & 6.09 & 6.26 & 0.44 \\
        LST noise scheduling & \textbf{2.20} & \textbf{2.09} & \textbf{0.70} \\
        \bottomrule
    \end{tabular}
    \caption{Ablation of the two types of noise scheduling.}
    \label{tab:results_ablation_noise_schedule}
\end{table}

%-------------------------------------------------------------------------
\subsection{Ablation study}

We investigate the impact of the noise scheduling strategy in the DiffTemp model,
and show the results in \cref{tab:results_ablation_noise_schedule}.
The results show a large performance gap between the pure noise scheduling and the LST noise scheduling.
This is expected as pure noise scheduling is for stochastic generation of images, 
where diversity weighs more than fidelity.
Though it fails to enforce a physical constraint on the generated $T_a$ images,
the perceptual performance measured by SSIM is still better than the ML methods.

%% file: sec/6_application.tex
\section{Potential applications}
\label{sec:application}

\begin{figure*}[htbp]
    \centering
    \includegraphics[width=\textwidth]{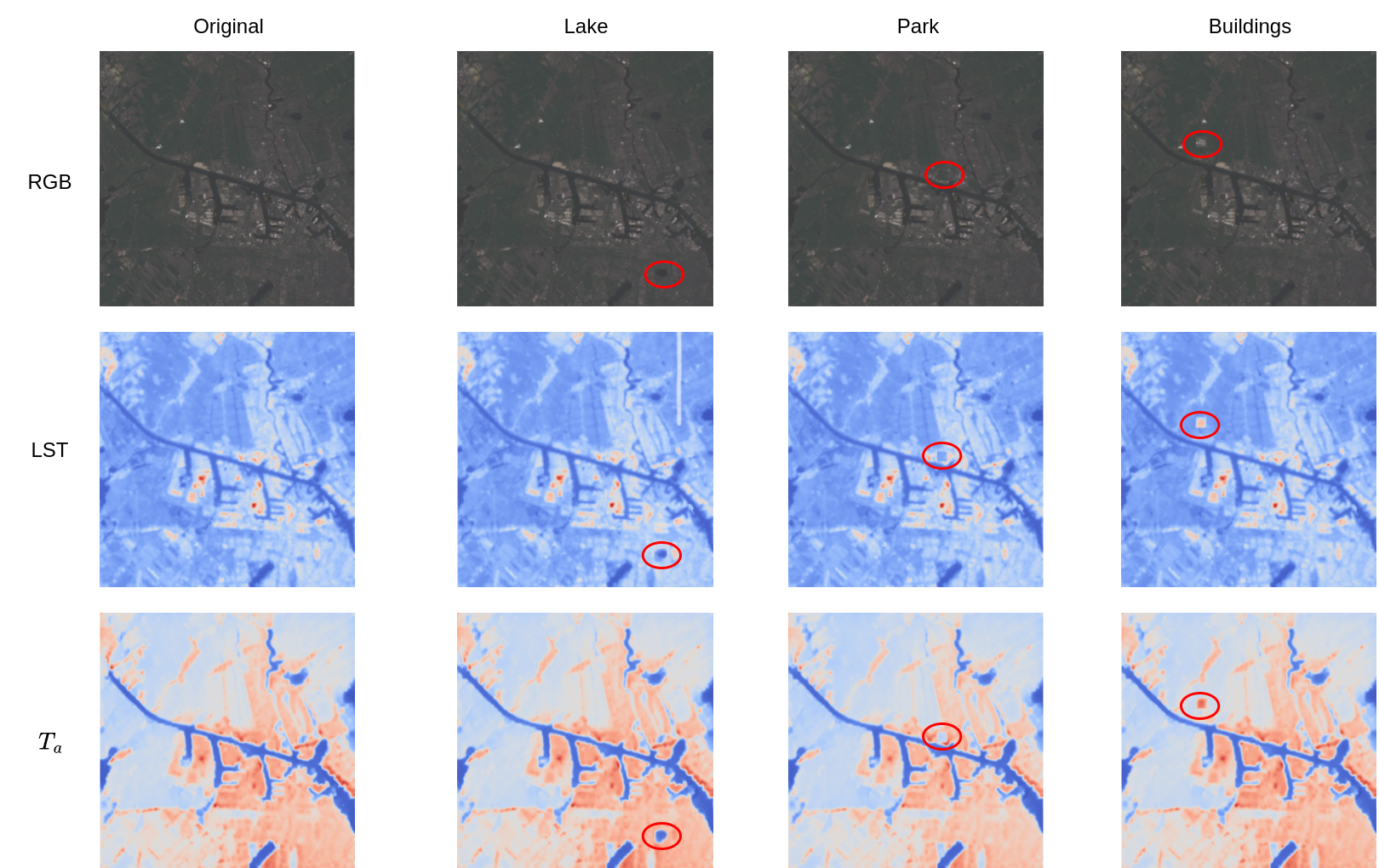}
    \caption{By simulating different urban designs, we show the impact of water bodies, green spaces, and buildings on the air temperature at neighborhood scale.
    Note: LST and $Ta$ are not drawn to the same scale for better visualizing the temperature characteristics within the image.}
    \label{fig:use_case}
\end{figure*}

%-------------------------------------------------------------------------

\textbf{Urban planning use case}. 
To illustrate the potential of DiffTemp in urban planning,
we simulate different urban designs by inserting different land features (denoted by $mask_{new}$) in the RGB images.
Due to the change of land features, the LST images should also change accordingly.
We build a model based on the same diffusion architecture to predict LST images from the RGB images alone.
Specifically, we use RGB as the condition and LST as the target image.
We use the noise scheduling strategy in \cref{eq:diffusion_forward_direct}.
At inference, given a modified RGB image with new land features, we generate new LST from the diffusion model.
With new RGB and LST inputs, we take them as the conditions and predict new $T_a$ using the trained DiffTemp model.

In particular, we simulate 3 types of urban designs:
1. insert water bodies (e.g. lake) in a vegetated area;
2. insert green spaces (e.g. park) in the residential area;
3. insert buildings (e.g. industry park) in the undeveloped area where vegetation is dominant.
Each newly inserted land feature is about $20 \times 20$ pixels which is equivalent to $600m \times 600m$ in the real world.
We illustrate the results in \cref{fig:use_case}.
It can be observed that the $T_a$ maps are significantly influenced by the new land features:
water bodies and green spaces tend to lower the $T_a$ while buildings tend to increase the $T_a$.
With this use case, we demonstrate the potential of DiffTemp in helping urban planners to assess the microclimate impact of different urban designs.

% -------------------------------------------------------------------------

\textbf{Other applications}
With the SR and Point SR task settings,
we enable prediction of HR $T_a$ maps from LR data sources and point observations respectively,
which is more practical in real-world scenarios.
With the predicted HR $T_a$, we further empower the study of UHI effect,
energy demand,
and human well-being.

%% file: sec/7_conclusion.tex
\section{Conclusion}

We propose a new benchmark for high-resolution (HR) air temperature ($T_a$) prediction, which is critical for urban planning, energy consumption, and human well-being.
Our dataset, LSTAT-20K, is the first to provide HR $T_a$ data with a large spatial area and extended temporal coverage.
We also are the first to leverage conditional diffusion models for this task
which has not been well studied from the CV perspective.
The extensive experiments demonstrate the effectiveness of our method in predicting $T_a$ not only for the physical accuracy but also for the perceptual quality.
We further present a use case of our method in urban planning,
which highlights an alternative approach other than costly numerical simulations for urban microclimate studies.